# Enhancing LLMs for Identifying and Prioritizing Important Medical Jargons from Electronic Health Record Notes Utilizing Data Augmentation: A Comparative Study

Won Seok Jang*, MSc[1], Sharmin Sultana*, BSc[1], Zonghai Yao*, MSc[2] Hieu Tran, MSc[2], Zhichao Yang, PhD[2], Sunjae Kwon, MSc[2], Hong Yu, PhD[1,3]

[1] Miner School of Computer and Information Sciences, UMass Lowell, MA, USA
[2] Manning College of Information and Computer Sciences, UMass Amherst, MA, USA
[3] Center for Healthcare Organization and Implementation Research, VA Bedford Health Care, MA, USA

These authors contributed equally *: Won Seok Jang, Sharmin Sultana and Zonghai Yao

Corresponding author: Hong Yu, Ph.D.

Phone: 1 978-934-3620 Email: Hong_Yu@uml.edu

**Abstract**

**Background**: OpenNotes allows patients to access their electronic health record (EHR) notes through online patient portals. However, EHR notes contain abundant medical jargon, which can be difficult for patients to comprehend. One way to improve comprehension is by reducing information overload and helping patients focus on the medical terms that matter most to them.

**Objective**: In this study, we evaluated both closed-source and open-source Large Language Models (LLMs) for extracting and prioritizing medical jargon from EHR notes relevant to individual patients, leveraging prompting techniques, fine-tuning and data augmentation.

**Methods**: We evaluated the performance of closed-source and open-source LLMs on a dataset of 90 expert-annotated EHR notes. We tested various combinations of settings including i) general and structured prompts, ii) zero-shot and few-shot prompting, iii) fine-tuning and iv) data augmentation. To enhance the extraction and prioritization capabilities of open-source models in low-resource settings, we applied data augmentation using GPT-4o and integrated a ranking technique to refine the training process. Additionally, to measure the impact of dataset size, we fine-tuned the models by incrementally increasing the size of the augmented dataset from 10 to 9,995 and tested their performance. The effectiveness of the models was assessed using 10-fold cross-validation, providing a comprehensive evaluation across various settings. We report the F1 score and Mean Reciprocal Rank (MRR) for performance evaluation using two different string matching algorithms (relaxed string matching and Jaccard Index). We also conducted error analysis classifying the erroneous outputs from the models.

**Results**: Our results show that open-source models achieved the highest performance, particularly when utilizing fine-tuning with a gold-standard dataset. Under Jaccard Index based string matching, DeepSeek 8B set the benchmarks with an F1 of 0.431 (SD 0.046); similarly, BioMistral 7B showed an MRR of 0.577 (SD 0.109). However, under relaxed string matching, open-source models were unable to match the performance of closed-source models, even with data augmentation or fine-tuning. We analyzed our experiment from several perspectives. First, few-shot prompting did not show an advantage over zero-shot prompting in vanilla models. Second, when comparing general and structured prompts, we found model performance could deviate largely based on prompting styles. Third, fine-tuning with a small gold-standard dataset improved performance. Lastly, data augmentation yielded performance comparable to or even surpassing the fine-tuning strategy. However, it also underscored the importance of the quality of the augmented dataset.

**Conclusion**: The evaluation of both closed-source and open-source LLMs highlighted the effectiveness of prompting strategies, fine-tuning, and data augmentation in enhancing model performance in low-resource scenarios.

# Introduction

## Background

Electronic Health Record (EHR) notes serve as valuable sources of information that can significantly benefit patients. Programs like OpenNotes [1] and the Blue Button [2] initiative empower patients by providing access to their EHR notes [3–9]. Nevertheless, the benefits of accessing EHR notes can diminish if patients do not comprehend their content [10–15]. EHR notes are lengthy and filled with medical jargon [16–19], which can be difficult to comprehend for the average U.S. adult, whose reading ability is around the 7th to 8th-grade level [20–24]. Therefore, supportive technologies are needed to assist patients in understanding EHR content [25,26], focusing on linking medical terms to lay-friendly terms [27–29], consumer-oriented definitions [18], and educational materials [30]. Early studies have demonstrated that such interventions significantly enhance patient understanding [18,27]. However, initial methods primarily relied on frequency- and context-based approaches to identify unfamiliar terms and propose simpler synonyms [27–29]. Identifying and extracting complex medical jargon from EHR notes is a crucial step toward improving patients' comprehension, ultimately enhancing patient engagement and reducing anxiety about their health [18,31–33].

Notably, not all medical jargon extracted from EHR notes holds equal clinical importance [34,35]. Existing tools, such as MetaMap [36], ScispaCy [37], medspaCy [38], and QuickUMLS [39], are effective at extracting medical terms—typically predefined terms from the Unified Medical Language System (UMLS)[40]—but often fail to prioritize these terms based on their relevance to individual patients, treating all terms as equally important [27–29]. In previous work, we asked physicians to identify medical jargon terms from EHR notes that are important to patients [34]. Our results showed that physicians were able to consistently identify 5 to 10 medical jargon terms from each EHR note and rank each term based on its importance to patients [34]. Furthermore, we developed feature-rich traditional machine learning models (e.g., support vector machines) to identify such terms [34]. However, our previous work did not focus on ranking jargon terms based on their importance to individual patients within an EHR note.

In this study, we propose large language model (LLM)-based NLP approaches to identify and rank jargon terms from EHR notes based on their importance to patients. LLMs have demonstrated tremendous promise in biomedical NLP applications [41–52] due to their exceptional generalizability and performance. However, applications in medical term extraction have primarily focused on tasks such as biomedical named entity recognition (BioNER) [53–55], rather than on prioritizing terms most relevant to patients, which is crucial for enhancing communication between patients and healthcare providers.

The key contributions of this paper are as follows:

- We conduct a comprehensive evaluation of both closed-source and open-source LLMs to assess their effectiveness in identifying medical jargon from EHR notes that are important for patients using a physician-annotated gold-standard dataset.
- We leverage data augmentation with AI-generated medical jargon from MIMIC-IV discharge summaries to address the challenges of training in low-resource settings. Under relaxed string matching, none of the methods surpassed the performance of closed-source models. In contrast, evaluation using the Jaccard Index showed that fine-tuning and data augmentation improved performance, exceeding that of closed-source models.

- We provide an in-depth analysis of the results from both quantitative and qualitative perspectives, focusing on common strategies for improving LLM performance, such as zero-shot and few-shot learning, prompt engineering, scaling laws, domain-adaptive training, and data augmentation, and recommendations for users based on our findings.

## Related Work

Identifying jargon terms important to patients is part of the biomedical Named Entity Recognition (BioNER) task, which involves identifying predefined entities in a text and labeling each token with the corresponding entity. Medical entities encompass categories such as diseases, medications, treatments, lab tests, and more [56,57]. Studies such as [58–61] have introduced language models for BioNER tasks, while more recent studies [53–55,62,63] have explored the application of large language models (LLMs) in BioNER. However, BioNER tasks primarily focus on extracting entities without considering their importance and relevance to the personal needs of patients, which distinguishes them from our objective.

MedJex [32] fine-tunes pre-trained language models (PLMs), such as BERT [64], RoBERTa [59], BioClinicalBERT [65], and BioBERT [58], on a domain-specific corpus. It leverages Wikipedia hyperlink spans during pretraining and transfers the learned weights to a target model fine-tuned on MedJ, an expert-annotated medical dataset. More recent studies [66] have investigated whether large language models (LLMs), such as ChatGPT [67], can outperform baseline PLMs (e.g., MedJEx [32] and SciSpacy [37]) in extracting personalized medical jargon. Similarly, GAMedX [68], a medical data extractor utilizing LLMs (Mistral 7B and Gemma 7B), employs chained prompts to navigate the complexities of specialized medical jargon. Other works [69–71] have demonstrated how LLMs can enhance the readability of EHR notes by extracting medical jargon.

This work also shares similarities with topic modeling, a task that extracts topics from input text. Using unsupervised learning algorithms, topic modeling can identify both explicit and implicit themes within a text corpus [72–74]. Through topic modeling, a text can be represented by multiple keywords or topics, which can then be incorporated into supervised models. However, topic modeling heavily relies on term frequencies and may easily overlook important terms that are clinically relevant to individual patients.

Among the most relevant works, such as FOCUS [34], ADS [75], and FIT [35], FOCUS [34] employs MetaMap [36] to extract medical jargon from EHR notes and utilizes feature-rich learning-to-rank techniques to determine whether the terms are important. However, none of the previous works have identified and ranked medical jargon terms in a note-specific manner. This is an important task, as ranking terms based on their relevance to a specific note may help the patient comprehend the note by linking important jargon terms to their lay definitions [76], or help generate patient-friendly after-visit summaries [77].

# Methods

## Overview

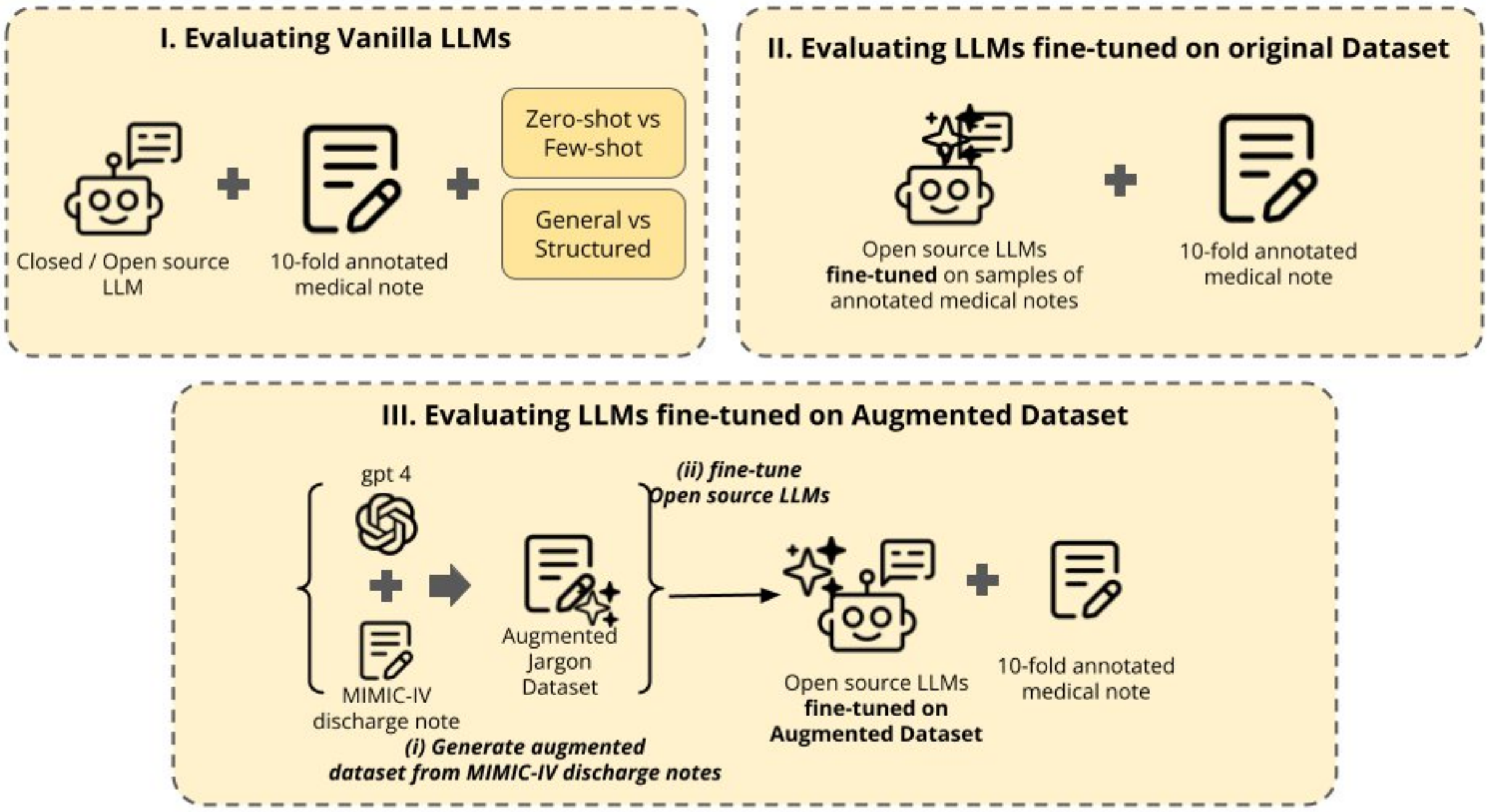

*Figure 1. The evaluation workflow for closed and open-source LLMs. We evaluate the performance of the LLMs in three distinctive settings. I. We assess the performance of closed- and open-source models by varying prompts and extraction tasks using the 10-fold annotated medical note (gold-standard dataset). II. Next, we benchmark the open-source models fine-tuned on portions of the Gold dataset. III. Finally, we apply data augmentation, generating synthetic data from GPT-4o to fine-tune the open-source LLMs and evaluate them under the same varying settings. Our study finds that using data augmentation, the models can reach comparable or even superior performance in personalized medical jargon extraction tasks.*

We evaluated both closed-source and open-source LLMs for their efficacy in extracting key information from annotated medical notes, aiming to assess performance across different strategies. Figure 1 provides an overview of our experiments, which leverage physician-annotated medical notes, closed- and open-source LLMs, and In-Context Learning (ICL). We examined the effects of prompting styles, fine-tuning, and data augmentation to enhance model performance.

## Data source

**An excerpt from Annotated EHR note**

Ms. XXX returns for follow-up of her rheumatoid arthritis [1] and a history of B-cell lymphoma. Since her last visit, Ms. XXX continues to have variable days of pain related to her postherpetic neuralgia [4]. She is frustrated by the persistent pain and wonders whether any of her medications for this is very helpful. She is interested in slowly tapering off her narcotics. As for her rheumatoid arthritis, she notes increasing symptoms in her hands and most recently, her knees and feet. She wonders if she should receive another dose of rituximab [2.1]. She just saw Dr. Y for follow-up of her large cell lymphoma [5]. She mentions that her labs revealed increased serum calcium, and she was told to hold her calcium supplements and vitamin D. As for her osteoporosis [3], she had an appointment with Dr. Z recently but then decided to get denosumab [2.2] here in our clinic. She is still concerned about the potential side effects of the medication. Ms. XXX's rheumatoid arthritis [1] symptoms have recently increased, particularly with the development of with synovitis [2], particularly in her left knee. The timing from her last dose of rituximab [2.1] is almost exactly a year. She and Dr. agree that it is reasonable to consider another infusion of this medication since she tolerated it well. The protocol would be for her RA rather than for her lymphoma. She is interested in receiving this through Dr. YY's office.

*Figure 2. A sample EHR note where physicians identified important medical terms. Diagnoses/conditions are highlighted in yellow, while medications, tests and procedures associated with those diagnoses are marked in green, accompanied by their respective rankings.*

Our gold-standard dataset consists of 90 medical notes, each annotated by two physicians [34,35]. For each note, agreement from both physicians was used as the final annotation. The annotation agreement (micro average) on these notes was 0.51 Cohen's Kappa [35]. This EHR note dataset comprises text reports across six medical categories: Cancer, COPD, Diabetes, Heart Failure, Hypertension, and Liver Failure (Table 1). Each medical note includes detailed patient information and is accompanied by physician annotations highlighting the most critical terms or phrases relevant to the patient's health status. Figure 2 presents a snippet of a sample EHR note from the gold-standard dataset.

| Main Diagnosis | Note Counts | Median Jargon Counts |
|---|---|---|
| Cancer | 17 | 8 |
| COPD | 20 | 8 |
| Diabetes | 15 | 8 |
| Hypertension | 11 | 6 |
| Liver Failure | 9 | 5 |
| Heart Failure | 18 | 7 |

*Table 1. Gold-Standard Dataset Description. The dataset consists of 90 notes from patients diagnosed with Cancer, Chronic Obstructive Pulmonary Disease (COPD), Diabetes, Hypertension, Liver Failure and Heart Failure. The median jargon counts for each category are illustrated.*

## Closed-Source and Open-Source LLMs

We used both publicly available LLMs (open-source LLMs) and proprietary models that are not publicly available (closed-source LLMs). The open-source LLMs included Mistral 7B v0.1 [78] ( Mistral 7B), BioMistral 7B [79], Llama3.1-8B [80] (Llama 3.1 8B), DeepSeek-R1-Llama-Distill-8B [81] (DeepSeek 8B). For proprietary models, we utilized two closed-source LLMs which were from OpenAI [67]: GPT-5.2, GPT-5-mini.

## Zero-shot vs Few-Shot

We used both zero-shot and few-shot prompts, also known as In-Context Learning (ICL), to compare the performance of the LLMs. For zero-shot prompting, we provided the model with general instructions, whereas for few-shot prompting, we included two examples randomly selected from the gold-standard dataset.

## General and Structured Prompts

To evaluate the model's performance, we implemented two variations of prompts: general prompts and structured prompts, each designed to assess how effectively the model could extract relevant clinical information from medical notes. Multimedia Appendix 1 presents an example of a structured prompt. In this approach, the prompts are explicitly designed to closely align with the original task defined in the gold-standard dataset. The structured prompts instruct the LLMs to extract key medical conditions or diagnoses, followed by the relevant medications associated. In contrast, general prompts provide a more flexible and broader context. These prompts instruct the model to extract key medical terms without explicitly differentiating between conditions and medications, assigning the same base rank to both. This approach allows the LLM to interpret the extraction task more broadly, offering insights into how well the model generalizes its understanding of medical terminology when provided with less specific guidance.

## Fine-tuning with Low-Rank Adaptation (LoRA)

To improve the performance of open-source LLMs (Mistral 7B, BioMistral 7B, Llama 3.1 8B and DeepSeek 8B), we conducted LoRA [82] based parameter-efficient finetuning (PEFT). LoRA is an efficient fine-tuning technique that allows models to be adapted to specific tasks without the need to update all of the model's parameters. Instead, it applies low-rank updates to specific layers, reducing the computational cost and memory usage typically associated with traditional fine-tuning methods. The training was done with a batch size of 1 per device and gradient accumulation over 128 steps, and low-rank dimension was set to 64. The learning rate was configured at $3e - 4$, and the models were trained over 5 epochs to allow the models to converge effectively on the task-specific patterns present in the dataset.

## Data Augmentation

Annotation by domain experts is expensive, and data augmentation using AI-generated data can help alleviate this challenge [83]. Few-shot prompting integrates task examples directly into input prompts, allowing models to observe patterns and generalize from limited examples to effectively handle new, unseen data [84]. We created an augmented dataset using the ICL technique, which was then used to refine the models (Multimedia Appendix 1). This augmented dataset was derived from discharge notes in the MIMIC-IV clinical database [85]. A subset of MIMIC-IV discharge notes was randomly selected, and GPT-4o from OpenAI [67] was used to process and rank key terms based on their importance for patient understanding. A dataset generated via GPT-4o was used because its outputs aligned more closely with the original annotations, exhibiting higher reliability during manual validation. The extraction process was guided by the examples, where two annotated notes from our gold-standard dataset were provided as examples to instruct the model on identifying and prioritizing terms.

After the raw generation, we filtered out terms that do not appear in the medical text using string matching. The resulting augmented notes were 9,995 notes. Using only a few-shot technique and no other filtering mechanism, we anticipated a lot of noise being injected in the augmented dataset. To this end, we further conducted an analysis on the augmented dataset, collecting 100 random cases and comparing the synthetic annotations with an expert annotation. We asked a clinician to annotate the MIMIC-IV notes with the same instructions given to the GPT-4o and compared the agreement between them.

## Exploring the Size of Augmented Dataset

To explore the effects of augmented dataset size, we progressively increased the dataset across four scales: 10, 100, 1,000, and 9,995 notes. By testing on the original gold-standard annotations along with AI-generated augmented data, we conducted a comprehensive evaluation of model performance across varying data sizes. Our objective was to enhance the robustness of LLMs in extracting critical medical information for patients.

## Baseline Models

We evaluated the performance of the open-source LLMs against several prominent baselines in the field of Named Entity Recognition (NER) tasks: MedJEx [32] and BioClinical-ModernBERT [86]. For MedJEx, we did not fine-tune the model since it is already a trained model to extract medical jargon. Also, MedJEx does not output any confidence score, therefore we could only report Precision, Recall and F1 score. For BioClinical-ModernBERT, we fine-tuned the model on our dataset and measured the performance. BioClinical-ModernBERT also outputs confidence scores. We used it as a proxy for ranking, assigning higher rankings for higher confidence scores. For the candidate entity extraction we used MedSpacy [38] and QuickUMLS [39] to form inputs for the baseline models.

## Evaluation Metrics

We used Precision (Eq. 1), Recall (Eq. 2), F1 (Eq. 3) and Mean Reciprocal Rank (MRR) (Eq. 4) as our metric. Performance metrics, including Precision, Recall, and F1-score, were first calculated at the individual note level ($i$). To aggregate these into a single representative score for each of the 10 folds, we calculated the macro-average across all notes within that fold. The final results reported are the mean and standard deviation of these fold-level averages. Precision and recall gave insights into the model's ability to cover the annotated terms. The Macro-F1 score allowed us to evaluate the model's balanced performance. MRR metric (Eq. 4), in particular, evaluated how well the model ranked the terms according to their relevance, reflecting the model's alignment with the annotated order of terms. Precision and Recall are provided in the Multimedia Appendix 2. For MRR calculation, integer valued ranks were used for both outputs from using general prompt and structured prompt. Because the structured prompt outputs used integer values in the MRR calculation, some results could receive tied ranks. To avoid inflating MRR, we used average ranks for tied results ($Adjusted\ Rank_i$). Specifically, when multiple terms shared the same rank, each term was assigned the average of the positions they occupied. This provides a more conservative evaluation by penalizing ambiguous predictions and rewarding models that rank the correct term more specifically. We used two different string matching approaches: 1) Relaxed string matching which checks the matching as true positives when it either perfectly matches or contains a gold-standard input [87,88], 2) The Jaccard Index method which uses set-based similarity to identify matches, employing Jaccard Distance to measure the gap between two strings. Here we set the similarity threshold of >0.5. Any extracted term achieving a Jaccard similarity score greater than 0.5 against the gold-standard term was counted as a true positive for the subsequent precision and recall calculations. Whereas Jaccard Index method penalizes verbosity, relaxed string matching does not. This enables us to have a clear picture about the extraction performance and quality.

$$Precision_i = \frac{TP_i}{TP_i + FP_i} \quad (1)$$

$$Recall_i = \frac{TP_i}{TP_i + FN_i} \quad (2)$$

$$F1_i = 2 \times \frac{Precision_i \cdot Recall_i}{Precsion_i + Recall_i} \quad (3)$$

$$MRR = \frac{1}{N}\sum_{i=1}^{N} RR_i \quad (4)$$

$$RR_i = \frac{1}{Adjusted\ Rank_i} = \left(\frac{1}{v}\sum_{j=k}^{k+v-1} j\right)^{-1} \quad (5)$$

Here, $RR_i$ denotes the Reciprocal Rank for each note $i$. $k$ is the starting rank or the tied rank, $v$ is the count of ties (Eq. 5). As shown in Eq. 5, we compute the arithmetic mean of the ordinal positions occupied by tied terms to determine the effective rank. The final Mean Reciprocal Rank ($MRR$) is then derived by averaging these adjusted $RR_i$ values across the total number of notes $N$ in the dataset (Eq. 4).

## Error Classification

To further understand the nature of the model's behavior, we conducted an error analysis of the models' outputs. First, we classified the model's outputs that were erroneous. We used three types of errors: (1) Spurious error (SE), where the model included terms that are not in the gold dataset but included in the medical note. (2) Granularity error (GE), which reflects overly specific or overly broad extractions, (3) Misalignment error (ME), where the model simply did not understand the instruction and misbehaved. Using this taxonomy, we identified the erroneous behavior of the LLMs.

## Experimental Details

All experiments were performed with two Nvidia A100 GPUs, each with 40 GB of memory, an Intel Xeon Gold 6230 CPU, and 192 GB of RAM. We used Python 3.9 and Huggingface transformers library [89] for our experiment. For the closed source models, we used OpenAI's ChatGPT API [67]. To reduce randomness of the experiment, we set the generation temperature to 0.1 for every model.

## Ethical Considerations

The requirement for ethical approval and informed consent was waived by the institutional review board at the VA Bedford Health Care System. The experiments were performed in accordance with the Declaration of Helsinki. The clinical data utilized for data augmentation were obtained from the MIMIC-IV database, a publicly available, de-identified repository of electronic health records from patients admitted to the emergency department or an intensive care unit at the Beth Israel Deaconess Medical Center in Boston, MA. Access to the database was granted following the completion of the required PhysioNet Credentialed Health Data Use Agreement 1.5.0.

# Results

| | F1 (relaxed) | F1 (Jaccard) | MRR (relaxed) | MRR (Jaccard) |
|---|---|---|---|---|
| MedJEx | 0.120 (0.027) | 0.138 (0.031) | - | - |
| BioClinical-Modern BERT | 0.076 (0.079) | 0.087 (0.086) | 0.087 (0.146) | 0.098 (0.151) |

*Table 2. Performance of Baseline models of F1 and MRR using relaxed string matching (relaxed) and Jaccard Distance based method (Jaccard)*

Table 2 presents the performance of the baseline models, which include commonly used language models (LMs) for medical information extraction. The highest baseline F1 score was achieved by MedJEx, 0.138 (SD 0.031) using Jaccard Index. For MRR, BioClinical-ModernBERT scored 0.087 (SD 0.146) and 0.098 (SD 0.151) using relaxed string matching and Jaccard Index respectively. Despite being fine-tuned on the dataset, BioClinical-ModernBERT did not show the highest performance. As we expected, the conventional methods have limitations in extracting medical entities that are particularly important to the patient. Given that these models have a parameter size of 150M — considerably smaller than the generative models used for comparison — this outcome is perhaps unsurprising.

| Models | Prompt | top5 | | | | top10 | | | |
|---|---|---|---|---|---|---|---|---|---|
| | | F1 (relaxed) | F1 (Jaccard) | MRR (relaxed) | MRR (Jaccard) | F1 (relaxed) | F1 (Jaccard) | MRR (relaxed) | MRR (Jaccard) |
| Closed-Source LLMs | | | | | | | | | |
| GPT-5.2 | general | **0.492 (0.068)** | 0.307 (0.071) | **0.578 (0.045)** | **0.476 (0.124)** | **0.496 (0.058)** | 0.31 (0.077) | **0.578 (0.045)** | **0.467 (0.117)** |
| | structured | 0.332 (0.077) | 0.206 (0.048) | 0.513 (0.153) | 0.359 (0.123) | 0.336 (0.073) | 0.211 (0.047) | 0.513 (0.153) | 0.359 (0.123) |
| GPT-5-mini | general | 0.46 (0.073) | 0.129 (0.048) | 0.561 (0.156) | 0.196 (0.11) | 0.505 (0.063) | 0.145 (0.045) | 0.552 (0.151) | 0.196 (0.11) |
| | structured | 0.387 (0.055) | 0.12 (0.046) | 0.575 (0.165) | 0.24 (0.17) | 0.387 (0.075) | 0.127 (0.052) | 0.566 (0.153) | 0.24 (0.17) |
| Open-Source LLMs | | | | | | | | | |
| Mistral 7B | general | 0.372 (0.085) | 0.323 (0.081) | 0.451 (0.175) | 0.453 (0.172) | 0.401 (0.119) | 0.345 (0.115) | 0.432 (0.167) | 0.444 (0.177) |
| | structured | 0.351 (0.083) | 0.243 (0.056) | 0.582 (0.108) | 0.455 (0.112) | 0.349 (0.081) | 0.236 (0.055) | 0.573 (0.106) | 0.455 (0.112) |
| Llama 3.1 8B | general | 0.347 (0.054) | 0.333 (0.055) | 0.426 (0.112) | 0.392 (0.108) | 0.368 (0.063) | 0.35 (0.056) | 0.422 (0.113) | 0.392 (0.108) |
| | structured | 0.371 (0.056) | 0.188 (0.051) | 0.521 (0.14) | 0.318 (0.112) | 0.358 (0.059) | 0.186 (0.055) | 0.517 (0.139) | 0.318 (0.112) |
| BioMistral 7B | general | 0.21 (0.076) | 0.017 (0.032) | 0.353 (0.141) | 0.033 (0.071) | 0.211 (0.078) | 0.02 (0.04) | 0.353 (0.141) | 0.033 (0.071) |
| | structured | 0.304 (0.121) | 0.017 (0.018) | 0.507 (0.152) | 0.056 (0.056) | 0.304 (0.123) | 0.017 (0.018) | 0.507 (0.152) | 0.056 (0.056) |
| DeepSeek 8B | general | 0.42 (0.077) | **0.383 (0.053)** | 0.416 (0.113) | 0.398 (0.13) | 0.443 (0.077) | **0.409 (0.068)** | 0.414 (0.113) | 0.397 (0.13) |
| | structured | 0.326 (0.026) | 0.211 (0.053) | 0.467 (0.131) | 0.4 (0.158) | 0.328 (0.048) | 0.218 (0.059) | 0.467 (0.131) | 0.4 (0.158) |

*Table 3. Comparison between closed/open-source vanilla models with zero-shot prompts. We compared the performance of different models using top-5, and top-10 results for F1 and MRR across both closed- and open-source LLMs, providing their mean and standard deviation. GPT-5.2 showed the highest performance of F1 and MRR using relaxed string matching, but DeepSeek 8B also showed the highest F1 score using Jaccard Index.*

| Models | Prompt | top5 | | | | top10 | | | |
|---|---|---|---|---|---|---|---|---|---|
| | | F1 (relaxed) | F1 (Jaccard) | MRR (relaxed) | MRR (Jaccard) | F1 (relaxed) | F1 (Jaccard) | MRR (relaxed) | MRR (Jaccard) |
| Closed-Source LLMs | | | | | | | | | |
| GPT-5.2 | general | **0.475 (0.068)** | 0.323 (0.06) | **0.561 (0.098)** | **0.536 (0.092)** | **0.478 (0.07)** | 0.326 (0.058) | **0.561 (0.098)** | **0.536 (0.092)** |
| | structured | 0.354 (0.069) | 0.233 (0.054) | 0.536 (0.174) | 0.387 (0.166) | 0.364 (0.076) | 0.237 (0.052) | 0.536 (0.173) | 0.387 (0.165) |
| GPT-5-mini | general | 0.475 (0.077) | 0.243 (0.087) | 0.543 (0.138) | 0.4 (0.133) | 0.492 (0.056) | 0.245 (0.069) | 0.543 (0.138) | 0.4 (0.133) |
| | structured | 0.348 (0.068) | 0.161 (0.041) | 0.508 (0.14) | 0.321 (0.172) | 0.347 (0.087) | 0.163 (0.052) | 0.495 (0.131) | 0.323 (0.174) |
| Open-Source LLMs | | | | | | | | | |
| Mistral 7B | general | 0.387 (0.046) | 0.338 (0.039) | 0.498 (0.141) | 0.503 (0.11) | 0.383 (0.05) | 0.335 (0.045) | 0.498 (0.141) | 0.503 (0.11) |
| | structured | 0.363 (0.067) | 0.335 (0.056) | 0.534 (0.104) | 0.533 (0.101) | 0.356 (0.073) | 0.332 (0.065) | 0.534 (0.104) | 0.533 (0.101) |
| Llama 3.1 8B | general | 0.372 (0.078) | 0.335 (0.093) | 0.521 (0.145) | 0.557 (0.126) | 0.369 (0.073) | 0.333 (0.096) | 0.521 (0.145) | 0.547 (0.132) |
| | structured | 0.368 (0.116) | **0.346 (0.111)** | 0.442 (0.066) | 0.502 (0.109) | 0.374 (0.126) | **0.353 (0.119)** | 0.442 (0.066) | 0.502 (0.109) |
| BioMistral 7B | general | 0.208 (0.063) | 0.08 (0.036) | 0.459 (0.115) | 0.3 (0.132) | 0.207 (0.067) | 0.081 (0.033) | 0.459 (0.115) | 0.3 (0.132) |
| | structured | 0.227 (0.085) | 0.098 (0.042) | 0.435 (0.072) | 0.318 (0.136) | 0.225 (0.084) | 0.097 (0.041) | 0.435 (0.072) | 0.318 (0.136) |
| DeepSeek 8B | general | 0.344 (0.058) | 0.335 (0.058) | 0.494 (0.125) | 0.485 (0.115) | 0.344 (0.066) | 0.334 (0.064) | 0.485 (0.116) | 0.476 (0.102) |
| | structured | 0.327 (0.105) | 0.299 (0.116) | 0.48 (0.173) | 0.466 (0.188) | 0.332 (0.105) | 0.304 (0.113) | 0.48 (0.173) | 0.466 (0.188) |

*Table 4. Comparison between closed/open-source vanilla models with few-shot prompts. We compared the performance of different models using top-5, and top-10 results for F1 and MRR across both closed- and open-*

*source LLMs, providing their mean and standard deviation. GPT-5.2 achieved the best F1, MRR score using relaxed string matching. However, Mistral 7B also achieved a better F1 score than GPT-5.2 using Jaccard Index.*

Table 3 and 4 present the results from both closed- and open-source models using zero-shot and few-shot prompts, respectively. The highest F1 score in zero-shot setting was 0.496 (SD 0.058), achieved by GPT-5.2. GPT-5.2 achieved the highest MRR in zero-shot prompts, 0.578 (SD 0.045) and 0.467 (SD 0.117) with relaxed string matching and Jaccard Index. Notably, DeepSeek 8B and Llama 3.1 8B achieved higher Jaccard Index scores than GPT-5.2. As this metric is inversely affected by verbosity—additional non-overlapping tokens increase the union while leaving the intersection unchanged—GPT-5.2's more expansive outputs likely explain its lower performance. We also observed a substantial gap between relaxed string matching and Jaccard Index scores across models. For example, BioMistral 7B (Tables 3 and 4) exhibited a notable discrepancy in F1 scores between these two metrics, indicating that vanilla models tended to produce verbose outputs.

There was no clear advantage of using a few-shot prompt. The highest scores were achieved from using zero-shot prompts. Although variations still exist, most of the models (GPT-5.2, GPT-5-mini, Mistral 7B, DeepSeek 8B) showed higher scores using general prompt over structured prompt. We hypothesize that each LLM has its own preferred prompts.

| Models | top 5 | | | | top 10 | | | |
|---|---|---|---|---|---|---|---|---|
| | F1 (relaxed) | F1 (Jaccard) | MRR (relaxed) | MRR (Jaccard) | F1 (relaxed) | F1 (Jaccard) | MRR (relaxed) | MRR (Jaccard) |
| Mistral 7B | 0.417 (0.093) | **0.43 (0.074)** | 0.525 (0.11) | 0.573 (0.111) | 0.416 (0.093) | 0.428 (0.074) | 0.525 (0.11) | 0.573 (0.111) |
| Llama 3.1 8B | 0.38 (0.065) | 0.384 (0.058) | 0.515 (0.078) | 0.527 (0.103) | 0.384 (0.07) | 0.388 (0.064) | 0.515 (0.078) | 0.527 (0.103) |
| BioMistral 7B | 0.374 (0.07) | 0.379 (0.062) | **0.565 (0.111)** | **0.577 (0.109)** | 0.377 (0.07) | 0.381 (0.061) | **0.565 (0.111)** | **0.577 (0.109)** |
| DeepSeek 8B | **0.42 (0.048)** | 0.429 (0.043) | 0.508 (0.085) | 0.501 (0.075) | **0.424 (0.052)** | **0.431 (0.046)** | 0.508 (0.085) | 0.501 (0.075) |

*Table 5. The average F1 and MRR score of fine-tuned open-source models with 10-fold Cross Validation. We used zeroshot prompting with general prompting style in top-5 and top-10 medical jargon extraction. In top 10 extraction task, DeepSeek 8B presented the highest F1 score in both metrics (relaxed, Jaccard). However, BioMistral 7B showed the highest MRR score in both metrics.*

Fine-tuning open-source models resulted in better performance compared to vanilla models. Here, we unified the prompt style to general prompt to evaluate the effectiveness of fine-tuning. The best performing model was DeepSeek 8B showing 0.424 (SD 0.052) in F1 (relaxed), 0.431 (SD 0.046) in F1 (Jaccard) in top 10 medical jargon extraction. For MRR, BioMistral 7B showed the best performance of 0.565 (SD 0.111) in MRR (relaxed) and 0.577 (SD 0.109) in MRR (Jaccard). Under relaxed string matching, fine-tuning did not outperform closed-source models such as GPT-5 (Table 3-5). However, it yielded notable improvements in domain-specific task performance. Conversely, fine-tuned models exceeded closed-source models under the Jaccard Index metric, indicating that fine-tuning reduced verbosity while maintaining relevant content in the generated outputs. Table 5 demonstrates a reduced margin between the Relaxed and Jaccard metrics.

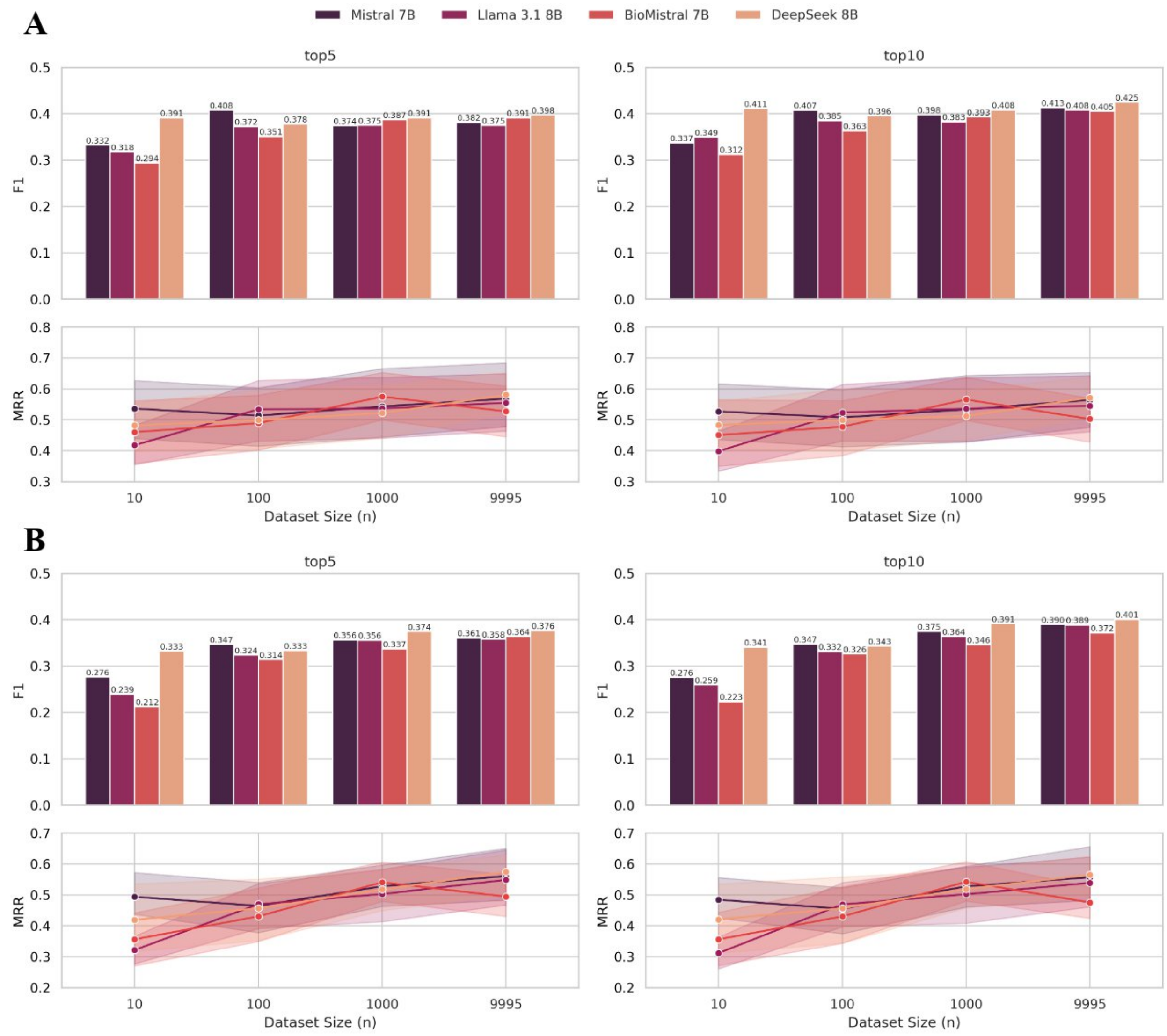

*Figure 3. Performance comparison of open-source LLMs across varying scales of MIMIC-IV augmented data. Evaluation is based on relaxed string metrics (A) and Jaccard distance (B), with training sizes ranging from 10 to 9995 samples. While individual model performance varies, a general trend of improvement is observed using Jaccard Index (B) as the dataset size increases. Bar graphs represent F1 scores, while line graphs depict Mean Reciprocal Rank (MRR). Both metrics indicate that DeepSeek 8B achieves the highest F1 score when trained on the largest dataset size, and the highest MRR.*

The highest-performing model that utilized data augmentation strategy was DeepSeek 8B, achieving an F1 score (relaxed) above 0.425 and F1 score (Jaccard) 0.401 in the top-10 medical jargon extractions (Figure 3, Multimedia Appendix 2. Supplementary Table 3A, 3B). The highest MRR score was achieved by DeepSeek 8B in the top-5, 10 medical jargon extractions using both metrics (relaxed, Jaccard). Performance patterns varied depending on the models and the metric. For example, under relaxed string matching, BioMistral 7B showed an increasing F1 trend, but MRR peaked at n=1,000 before declining in both top-5 and top-10 extraction tasks. In contrast, Mistral 7B achieved peak performance at n=100 for top-5 extraction, while DeepSeek performed best at n=10 and n=9,995. However, when evaluated using the Jaccard Index, all models exhibited consistently increasing F1 scores as dataset size increased. These findings are discussed further below. Although the results failed to achieve higher scores than closed-source models under relaxed string matching, utilizing the augmented dataset was found useful in some models, such as BioMistral 7B, Llama 3.1 8B and DeepSeek 8B, but not in Mistral 7B. However, under Jaccard Index, our results show that data augmentation could also outperform closed-source models.¶

| Model versions \ Errors | Spurious Error (n) | Granularity Error (n) | Misalignment Error (n) | Total Errors Count (n) |
|---|---|---|---|---|
| Vanilla model | 76 | 7 | 37 | 120 |
| Fine-tuned | 46 | 1 | 33 | 80 |
| Augmented Dataset | 80 | 2 | 33 | 115 |

*Table 6. The error analysis table. We suggest three types of errors in the system: Spurious Error, Granularity Error and Misalignment Error. All models show a high level of Spurious Error but fine-tuning substantially reduces the error rate, showing the lowest total error counts among other methods.*

To further investigate error distributions, we evaluated the performance of the vanilla DeepSeek model against its fine-tuned and augmented counterparts (n=9,995) using a representative test set from our 10-fold cross-validation. As illustrated in Table 6, the vanilla DeepSeek model demonstrated the highest frequency of Misalignment and Granularity Errors. While fine-tuning substantially reduced all error types—most notably Spurious Errors, which dropped to 46—the total error count was lowest for the fine-tuned model. Conversely, the data augmentation strategy did not yield similar improvements; it resulted in 80 Spurious Errors—surpassing the vanilla model—and only marginal reductions in Granularity and Misalignment Errors. Ultimately, the data augmentation approach failed to achieve the high-magnitude error reduction observed with standard fine-tuning.

# Discussion

## Zero-shot vs. Few-shot

| True Labels | Mistral 7B zero-shot | Mistral 7B few-shot |
|---|---|---|
| 1. Syncope | 1. syncope | 1. COPD |
| 2. Cough-induced syncope | 2. cough | 1. sleep apnea |
| 4. Arrhythmia | 3. near syncope | 1. coronary artery disease |
| 4. Dysrhythmias | 4. vertigo | 1. GERD |
| 7. Autonomic dysfunction | 5. falls | 1. hypertension |
| 8. Benign positional vertigo | 6. head trauma | 1. osteoarthritis |
| 8. Vertigo | 7. fracture | 1. carpal tunnel syndrome |
| | 8. dysrhythmias | 1. nephrolithiasis |
| | 9. arrhythmia | 1. degenerative disc disease |
| | 10. EEG | 1. depression |
| | 11. MRI | 1. erectile dysfunction |
| | 12. brain | 1. adenomatous polyps of the colon |
| | 13. episodes | 2. Losartan |
| | 14. cough-induced syncope | … |
| | 15. tonic-clonic activity | 2. Iron sulfate |
| | … | 2. Tylenol |
| | 26. soft tissue attenuation | 2. Oxycodone |
| | 27. cardiac catheterization | 2. Aspirin |
| | 28. left main | 2. Omeprazole |
| | 29. left circumflex | 2. Trileptal |
| | 30. proximal right coron | 2. Flomax |

*Figure 4. Case Study for extracting top 10 important medical jargons from Mistral 7B on zero-shot and few-shot settings. Interestingly, few-shot prompting fails to extract terms that clinicians annotated as "gold", whereas zero-shot successfully conducts the task.*

Our findings suggest that there are minimal differences between zero-shot and few-shot prompting. Although there are some cases where a few-shot prompt outperformed a zero-shot prompt, these results are not always consistent. Results differ by model and it is hard to tell which prompting — zero-shot or few-shot — performs better. This is consistent with [90], which found that few-shot prompting is not always more advantageous than zero-shot prompting.

## Prompting Styles

The effectiveness of prompts can vary across models, with certain prompt styles enhancing performance for specific models (Table 3). While differences between models were generally minimal, some models performed better with particular prompt styles. For example, in the Llama 3.1 8B model, structured prompts outperformed general prompts with few-shot prompting. Whereas in Mistral 7B and in other models, general prompts showed improved performance over structured prompts regardless of string matching metrics. For both Llama 3.1 8B and Mistral 7B, structured prompts induced higher recall than precision. Since structured format gives more specific instruction, this may cause the model to generate concise and precise results (Multimedia Appendix 2. Supplementary Table 1A,1B). A plausible explanation is that structured prompts impose stronger output constraints, narrowing the model's generation space. Largely, our results align with prior research suggesting that tailored prompts can optimize a model's performance in specialized tasks [91]. Interestingly, in most cases, we found F1 score trends are aligned with MRR scores, suggesting that certain prompts helped the model reach its full potential. This shows that testing diverse prompt styles is essential in maximizing model performance, which is highly aligned with the current research results [92–95].

## Fine-tuning with Gold-Standard Data

| True Labels | BioMistral 7B zero-shot | BioMistral 7B fine-tuned |
|---|---|---|
| 1. Aspirin | 1. Coronary artery disease. No recurrent chest pain since that isolated event on XXXX. He remains on aspirin, beta-blocker and statin therapy. | 1. Aspirin |
| 1. Atorvastatin | | 1. Atorvastatin |
| 1. Echocardiogram | | 1. Metoprolol |
| 3. Abilify | | 1. Plavix |
| 3. COPD | | 1. sleep apnea |
| 3. Carbamazepine | | 1. weight gain |
| 3. Flovent | | 2. Carbamazepine |
| 3. Hypertension | | 2. Lithium carbonate |
| 3. Lithium | | 2. schizophrenia |
| 3. ProAir | | |
| 3. Schizophrenia | | |
| 4. Hypertriglyceridemia | | |
| 5. Sleep apnea | | |

*Figure 5. Case Study for extracting Top 10 important medical jargons from BioMistral 7B and fine-tuned BioMistral 7B on the gold-standard dataset. The fine-tuned model shows more robustness than vanilla models. The highlighted jargons are the ones that overlap with the expert-annotated labels, whereas BioMistral 7B shows Misalignment Erroneous output. After a few steps of fine-tuning, the model's behavior was significantly adjusted. Additionally, the verbosity of the output has decreased.*

Fine-tuning LLMs using domain-specific data proved effective in enhancing model performance, even surpassing the performance of closed-source models (GPT-5) under Jaccard Index (Table 5). The size of the datasets used for fine-tuning was small but led to performance gains. Increasing the size of the fine-tuning dataset is likely to further improve the model's performance [96]. Figure 5 illustrates the impact of fine-tuning: while vanilla models show limitations in extracting key points using prompts alone, fine-tuning enables the models to learn relevant patterns, leading to improved performance. Although BioMistral 7B could extract some information with instructions, MEs were present in its output. After conducting fine-tuning, the model dramatically improved performance, reducing its errors as shown in Figure 5. This trend is further evidenced by the DeepSeek 8B results in Table 6, where the SE exhibits a substantial reduction. Furthermore, the verbosity of model outputs was significantly reduced after fine-tuning, as reflected in improved Jaccard Index scores and overall performance gains. Previous research has similarly shown that fine-tuning on domain-specific data can significantly enhance performance by adjusting model weights to reflect the unique characteristics of the target domain, such as better handling of abbreviations, acronyms, and clinically relevant contexts [58]. Moreover, instruction fine-tuning has been shown to improve the model's zero-shot performance [97].

Our results reveal that different string matching metrics yield divergent performance narratives. Closed-source models set the benchmark under relaxed string matching; however, fine-tuned models achieved superior F1 scores and MRR when evaluated using the Jaccard Index. We contend that employing a combination of string matching metrics enables a more nuanced assessment of model performance. Specifically, relaxed string matching prioritizes semantic preservation and the recognition of domain-specific terminology, while the Jaccard Index measures exact token-level overlap, thereby penalizing verbosity and rewarding concise outputs.

## Analysis of Augmented Dataset

The data augmentation process relied solely on few-shot prompting, with string matching-based filtering as the only quality control measure. Consequently, the augmented dataset may be of lower quality than the gold-standard. To evaluate this, we randomly sampled 100 instances from the augmented dataset and had a clinical expert manually annotate them. Agreement between the expert's annotations and the synthetic labels was measured using the Jaccard Index [98], and to measure the ranking agreement, MRR. The scores were 0.255 and 0.585, respectively. The terms extracted showed low Jaccard distance, which likely contributed to lower performance. However, for MRR, the ranking showed comparable performance.

## Fine-Tuning with Augmented Data

We explored the impact of data augmentation using various sizes of MIMIC-IV [85] discharge notes to enhance the performance of LLMs. Data augmentation simulated diverse scenarios within medical notes, enabling the LLM to generalize better across a broader range of note types. In accordance with fine-tuning, the performance gains from data augmentation were meaningful (Figure 3). In some models, we found utilizing augmented data was helpful in improving the performance (Llama 3.1 8B, BioMistral 7B, DeepSeek 8B) under relaxed string matching in the top10 extraction task. Furthermore, all models trained on the maximum dataset size (n=9,995) demonstrated improved F1 scores that exceeded those of closed-source models under the Jaccard Index based string matching, but not under relaxed string matching. These findings align with prior studies [97,99], which argue that using LLM-generated data can improve model performance on downstream tasks. Additionally, studies have shown that while data augmentation provides benefits, substantial improvements often require a high degree of variation in the augmented data [100,101].

However, fine-tuning with augmented dataset did not outperform fine-tuning with the gold-standard dataset for other models (Mistral 7B, BioMistral 7B, DeepSeek 8B) when evaluated using Jaccard Index-based string matching (Table 5). Since we only used two examples for ICL, this may have affected the overall quality of the augmented dataset [102]. Nonetheless, this still leaves a question of why the LLMs' performance increased in the first place, considering the quality of the augmented dataset. Although data augmentation did not generate the most "accurate" synthetic dataset, it did not generate critical errors that severely harmed the model's performance. As a result, the augmented data provides weak but structured supervision that remains correlated with expert judgment. Consistent with prior findings in weakly supervised learning, such signals can still guide models toward meaningful patterns, particularly when aggregated across large-scale data. The observed performance improvements across models suggest that the augmented dataset captures clinically relevant structure despite instance-level disagreement, supporting the validity of our approach.

## Impact of Augmented Dataset Size

The effect of augmented dataset size on model performance varied depending on the string matching metric. Under relaxed string matching, only BioMistral 7B demonstrated a consistent performance improvement with increasing dataset size (Figure 3A). Conversely, string matching using the Jaccard Index showed that all models achieved higher F1 scores as the dataset size grew (Figure 3B), indicating metric-dependent performance trends. Our results partially align with findings from [103,104], where increasing the size of synthetic datasets significantly enhanced model performance compared to models trained on smaller real-world datasets. For other models such as DeepSeek 8B, the model's performance oscillated over dataset size in top5 and top10 extraction under relaxed string matching, although it ultimately showed the best performance at n=9,995 regardless of the metrics.

## Recommendations for Best Practices

Based on our findings, we categorize the risks of deploying LLMs into two distinct domains: Operational Risk and Clinical Risk. (1) Operational Risk encompasses failures in prompt adherence or structural inconsistencies where the model's output deviates from the required format, potentially causing system-level failures. These risks are largely mitigable through rule-based validation and explicit format-checking mechanisms. (2) Clinical Risk refers to the generation of factually incorrect information (hallucinations) that could directly compromise patient safety. Our results demonstrate that despite performance improvements across models, clinical risk remains a critical concern. Consequently, we argue that all LLM-generated outputs must undergo human-in-the-loop verification and be subjected to rigorous fact-checking protocols before clinical implementation.

## Limitations

This study has several limitations. First, we tested only a limited selection of available closed- and open-source LLMs. Specifically, we only tested models with fewer than 10B parameters, which are relatively small open-source LLMs. Second, our gold-standard dataset, being based solely on physician annotations, captures only the clinical perspective, not considering the actual needs of patients. Third, despite the improvements achieved through fine-tuning and data augmentation, the performance of the models still falls short of human annotation. Fourth, the effectiveness of these methods should be validated in real-world settings, such as Hyper-DREAM which tests the efficacy of the blood pressure management system [105].

# Conclusion

Our study provides a comprehensive evaluation of closed- and open-source LLMs for identifying and prioritizing medical jargon within expert-annotated EHR notes. Strategic interventions—prompting, fine-tuning, and data augmentation—notably enhanced the utility of open-source models for domain-specific clinical tasks. Fine-tuning on a gold-standard dataset emerged as the most effective overall strategy, yielding the highest performance gains and a substantial reduction in error counts under Jaccard Index based string matching. While the data augmentation strategy did not consistently outperform fine-tuning, it demonstrated measurable improvements in model-dependent scenarios. Nevertheless, our error analysis underscores that all evaluated models, regardless of architecture or optimization status, retain inherent degrees of both operational and clinical risk.

# Acknowledgments

We greatly value UMass BioNLP group's insightful feedback and thoughtful guidance. All the sections were revised through generative AI tools for checking grammatical errors and correcting awkward sentences. The views expressed in this article are those of the authors and do not necessarily reflect the position or policy of the United States Department of Veterans Affairs or the United States government.

## Author Contributions

Won Seok Jang, Sharmin Sultana, and Zonghai Yao authored the manuscript, with Won Seok and Sharmin implementing and analyzing models and Zonghai contributing to project planning.

Zhichao Yang, Hieu Tran, and Sunjae Kwon designed the framework and reviewed outcomes.

Hong Yu planned the project, wrote the proposal, and guided the research.

## Conflict of Interest Statement

No conflicting interests.

## DATA AVAILABILITY

The source code will be released here : https://www.github.com/memy85/2024_medicalnote_annotation

## Multimedia Appendix 1. Prompts used for experiments

## Multimedia Appendix 2. Supplementary results with Mean and Standard Deviation of Precision, Recall

¶